\documentclass[letterpaper, 10 pt, conference]{ieeeconf}
\IEEEoverridecommandlockouts

\pdfoutput=1
\usepackage{url}

\usepackage{graphicx}
\usepackage{amsmath,amssymb,amsfonts}
\usepackage{hyperref}
\usepackage{bm}
\usepackage{xcolor}
\usepackage{multirow}
\usepackage{paralist}
\usepackage{wrapfig}
\usepackage[capitalise]{cleveref}
\usepackage{siunitx}
\usepackage{booktabs}
\usepackage{caption}
\usepackage{graphicx}
\usepackage{etoolbox,siunitx,booktabs,threeparttable,multirow,array,graphicx}
\usepackage[normalem]{ulem}
\usepackage{makecell}
\usepackage{siunitx}
\usepackage{pifont}
\robustify\bfseries
\robustify\uline
\robustify\tnote
\usepackage[utf8]{inputenc}
\usepackage{pgf}
\usepackage{float}
\usepackage[super]{nth}
\DeclareUnicodeCharacter{2212}{−}

\usepackage[style=ieee,citestyle=numeric-comp]{biblatex}
\usepackage[ruled,linesnumbered]{algorithm2e}
\usepackage{placeins}
\usepackage{tikz}
\usepackage{pgfplots}

\usepackage[table]{xcolor}
\definecolor{rankone}{HTML}{F5E6AD}
\definecolor{ranktwo}{HTML}{F39192}
\definecolor{rankthree}{HTML}{F13C77}

\usepackage[nolist,nohyperlinks]{acronym}
\begin{acronym}
\acro{AbsRel}{Absolute Relative Error}
\acro{BT}{Ballast tank}
\acro{CFAR}{constant false alarm rate}
\acro{CNN}{Convolutional Neural Network}
\acro{CRF}{Conditional Random Field}
\acro{DAV2}{Depth Anything V2}
\acro{DAV2}{Depth Anything V2}
\acro{DC}{depth completion}
\acro{EDF}{Euclidean Distance Field}
\acro{ESDF}{Euclidian Signed Distance Field}
\acro{FF}{Farm field}
\acro{FMCW}{frequency-modulated continuous-wave}
\acro{GF}{Glacier flight}
\acro{GNSS}{Global Navigation Satellite System}
\acro{IH}{Industrial hall}
\acro{IMU}{Inertial Measurement Unit}
\acro{LiDAR}{Light Detection And Ranging}
\acro{LS}{Least Squares}
\acro{MAE}{Mean Absolute Error}
\acro{MAV}{Micro Aerial Vehicle}
\acro{MDE}{monocular depth estimation}
\acro{MSE}{Mean Squared Error}
\acro{MVS}{Multi-View Stereo}
\acro{NLL}{Negative Log-Likelihood}
\acro{OMAV}{Omnidirectional Micro Aerial Vehicle}
\acro{RMP}{Riemannian Motion Policy}
\acro{RMSE}{Root Mean Square Error}
\acro{SD}{sparse depth}
\acro{SfM}{Structure from Motion}
\acro{SLAM}{Simultaneous Localization and Mapping}
\acro{SNR}{signal-to-noise ratio}
\acro{SOTA}{state-of-the-art}
\acro{ToF}{Time of Flight}
\acro{UAV}{Uncrewed Aerial Vehicle}
\acro{UF}{Underwater fjord}
\acro{ViT}{Vision Transformer}
\acro{VTOL}{vertical take-off and landing}
\acro{DPT}{dense prediction transformer}
\newacroindefinite{IMU}{an}{an}
\newacroindefinite{MAV}{an}{a}
\newacroindefinite{OMAV}{an}{an}
\end{acronym}

\begin{document}
\newcommand{\todo}[1]{}
\renewcommand{\todo}[1]{{\color{red}\uline{TODO: #1}}}

\title{Depth Completion in Unseen Field Robotics Environments Using Extremely Sparse Depth Measurements}
\author{Marco~Job,
        Thomas~Stastny,
        Eleni~Kelasidi,
        Roland~Siegwart,
        Michael~Pantic
    \thanks{
        The authors MJ, TS, RS and MP are with the Autonomous Systems Lab, ETH Z\"urich, Switzerland.
        The corresponding author MJ ({\tt\footnotesize marco.job@ntnu.no}) and EK are with the Field Robotics Lab, NTNU, Norway.
        This work has been supported by a Swiss Polar Institute Technogrant, the Armasuisse Research Grant No 4780002580, and by an ETH RobotX research grant funded through the ETH Zurich Foundation.
        The code and datasets are available at: {\tt\footnotesize github.com/ntnu-frl/depth-completion}.
    }
}

\maketitle
\thispagestyle{empty}
\pagestyle{empty}
\begin{abstract}
Autonomous field robots operating in unstructured environments require robust perception to ensure safe and reliable operations.
Recent advances in monocular depth estimation have demonstrated the potential of low-cost cameras as depth sensors; however, their adoption in field robotics remains limited due to the absence of reliable scale cues, ambiguous or low-texture conditions, and the scarcity of large-scale datasets.
To address these challenges, we propose a depth completion model that trains on synthetic data and uses extremely sparse measurements from depth sensors to predict dense metric depth in unseen field robotics environments.
A synthetic dataset generation pipeline tailored to field robotics enables the creation of multiple realistic datasets for training purposes.
This dataset generation approach utilizes textured 3D meshes from Structure from Motion and photorealistic rendering with novel viewpoint synthesis to simulate diverse field robotics scenarios.
Our approach achieves an end-to-end latency of 53\,ms per frame on a Nvidia Jetson AGX Orin, enabling real-time deployment on embedded platforms.
Extensive evaluation demonstrates competitive performance across diverse real-world field robotics scenarios.
\end{abstract}

\section{Introduction}
Performing autonomous operations in challenging field robotic applications requires a comprehensive understanding of the surroundings, particularly in unknown and unstructured environments.
This understanding improves the efficiency, safety, and effectiveness of operations, supporting the development of robots capable of performing complex tasks in diverse and demanding settings.
Its importance is particularly evident in navigation, where robots must perceive the geometric properties of the scene.
Beyond navigation, it is crucial for obstacle avoidance, mapping, and informed decision making, all of which are key enabling factors to ensure autonomy and reliability in field scenarios.
To fulfill these requirements, robots commonly employ depth sensors from a range of existing modalities, such as \acp{LiDAR}, \ac{ToF} cameras and sensors, radars, and stereo cameras \cite{cadena2016past,nissov2024degradation}.
\begin{figure}[ht!]
    \centering
    \includegraphics[width=\columnwidth]{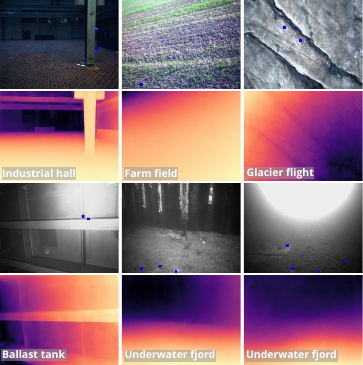}
    \caption{Our \acf{DC} approach in five unseen, real-world field robotics environments. The input is a three-channel image and a fourth channel for extremely sparse depth measurements. The blue squares visualize the measurements in this figure. The output is dense metric depth.}
    \label{fig:introduction}
\end{figure}
Recent advances in metric \ac{MDE} have motivated the use of simple and inexpensive monocular cameras as depth sensors \cite{ke2024repurposing,hu2024metric3dv2,yang2024depth_anything_v2,piccinelli2025unik3d,bochkovskii2025depthpro}.
These works have focused on generalization capabilities in structured domains, utilizing large-scale datasets to build robust models.
In general, \ac{MDE} approaches estimate metric depth using scale cues present in the images, as the problem is inherently ill-posed \cite{ke2024repurposing}.

However, the direct applicability of \ac{MDE} methods for metric depth estimation in real-world field robotics environments remains challenging for vision-based approaches due to the frequent absence of scale features and the prevalence of self-similar, ambiguous, or low-texture conditions \cite{cadena2016past,nissov2024degradation} (see \cref{fig:introduction}).
In addition, few large-scale datasets dedicated to field robotics exist, reflecting the inherent difficulty of collecting data under such challenging conditions.
A potential solution to overcome the limitations of metric \ac{MDE} is to use additional depth measurements to disambiguate the scene.
This approach, known as \acf{DC}, commonly uses a set of sparse depth measurements and fills in the gaps to obtain a dense depth image \cite{uhrig2017sparsity}.
Depending on the specific constraints of the application, domain or robotic platform, the order of magnitude of depth points can be as low as $10^{1}$ per frame.
This is common with commercially available, low-cost, and small-size sensors, such as \ac{FMCW} radars, single- or few-point \ac{ToF} sensors, and visually tracked sparse 3D landmarks.
Moreover, \acp{LiDAR} can also produce sparse and noisy depth measurements if the sensor operates in a degraded mode \cite{nissov2024degradation}.

Few works have explored the use of such sparse, noisy, and degraded measurements for \ac{DC} \cite{ma2018sparsetodense,qiu2019deeplidar,lin2020depth,long2021radar,feng2022advancing}.
Even fewer studies have investigated this sparsity level combined with generalization to unseen real-world field robotics environments \cite{wang2024g2monodepth,viola2025marigolddc,zuo2025omni,park2024depthprompting}.
To this end, motivated by this research gap and the lack of field robotics-focused datasets, the main contributions of this paper are summarized as follows:
\begin{itemize}
    \item A real-time \acf{DC} approach that predicts metric dense depth from input images and extremely sparse depth measurements (see \cref{fig:introduction}). Our approach architecturally extends a \ac{SOTA} \ac{MDE} method, incorporating the sparse depth measurements, while retaining the \ac{MDE}'s generalization capabilities.
    \item A synthetic training dataset generation approach tailored to field robotics that utilizes 3D meshes obtained from \ac{SfM}, photorealistic rendering, and novel viewpoint synthesis.
    \item The release of the four synthetic training datasets, code, trained model, and real-world evaluation datasets to support future research in this domain.
\end{itemize}

\section{Related Work}
Eigen et al. \cite{eigen2014depth} strongly influenced the field of \ac{MDE} with one of the first end-to-end metric depth estimation networks.
Since then, the field has seen numerous works \cite{laina2016deeper, liu2016learning, fu2018dorn,yang2021transformer,bhat2021adabins,saxena2023surprising}, introducing various architectures and performance improvements.
Yet, an inhibitor for wide-scale adaptation in field robotics is the lack of generalization to domains outside the training set.

The research challenge of generalizing to unseen environments is known as zero-shot depth estimation \cite{ranftl2022towards}.
In this domain, considerable research efforts have focused on training with large-scale datasets and affine-invariant depth estimation \cite{ranftl2022towards,ke2024repurposing,yang2024depth_anything_v2,wang2025moge}.
For field robotics, metric depth estimation is more crucial, and recent works have shown increasing performance primarily in structured environments \cite{bhat2023zoedepth,guizilini2023towards,hu2024metric3dv2,bochkovskii2025depthpro,piccinelli2025unik3d}.
Notably, \cite{guizilini2023towards} encodes camera intrinsics in the input space and \cite{hu2024metric3dv2} employs a canonical image transformation, both to improve generalization with varying cameras.
The authors of \cite{piccinelli2025unik3d} propose a method that generalizes to a wide range of camera models and intrinsics.
In the absence of scale cues, these methods still attempt to solve an ill-posed problem, which is an inherent limitation of \ac{MDE} rather than the models themselves \cite{ke2024repurposing}, motivating the incorporation of additional depth information.

To overcome metric \ac{MDE} limitations, \acf{DC} methods have been investigated.
Initial \ac{DC} works utilized only \ac{LiDAR} scans and filled in the gaps in the depth image \cite{uhrig2017sparsity}.
Later works began incorporating RGB images to improve the performance.
These methods often utilize \ac{LiDAR} or automotive radar scans, or randomly subsample the ground truth to simulate a sparse sensor \cite{ma2018sparsetodense,qiu2019deeplidar,lin2020depth,long2021radar,feng2022advancing}.
Nevertheless, these works are typically trained on constrained domains and thus demonstrate restricted applicability to unseen environments outside the training set \cite{viola2025marigolddc,zuo2025omni}.

Recently, a direction of generalizable \ac{DC} has emerged, with notable works including \cite{wang2024g2monodepth,viola2025marigolddc,zuo2025omni,park2024depthprompting}.
The authors of \cite{wang2024g2monodepth} propose a depth data augmentation framework, and a U-Net-based architecture for metric depth prediction.
The approach in \cite{zuo2025omni} uses a custom layer that solves a linear \ac{LS} problem to incorporate multi-resolution depth gradients and the sparse depth measurements as constraints into a dense depth map.
In \cite{park2024depthprompting}, the authors utilize a depth foundation model with a prompting module to guide the prediction to the metric domain.
These works still utilize approximately one order of magnitude more sparse depth points than what can be available in challenging field robotics environments.
Furthermore, since the focus is not field robotics, the training and evaluation datasets in these works are often not representative of this domain.
Notably, \cite{viola2025marigolddc} proposes a diffusion-based \ac{DC} approach that combines a pretrained affine-invariant \ac{MDE} model with very sparse depth measurements for metric depth.
Due to the diffusion-based architecture, real-time capabilities are a concern, which is particularly relevant for field robotics applications.

Another challenge for increasing adoption in the field robotics domain is the limited availability of training datasets.
Several depth datasets exist for \ac{MDE} and \ac{DC}, focused on automotive, indoor and outdoor \cite{silberman2012indoor,uhrig2017sparsity,vasiljevic2019diode} environments.
These datasets often contain numerous scale cues, such as vehicles, furniture, or man-made structures, and do not cover the diverse range of natural environments in the field.
Therefore, these widely used datasets may not be fully representative of challenging field robotic environments.

The acquisition of large-scale data in-the-wild is extremely labor-intensive and generally challenging.
The work in \cite{li2018megadepth} is notable for employing \ac{SfM} to scale up the collection of metric depth maps from publicly available images.
However, due to the limited availability of field robotic datasets, these environments are scarcely represented in this pool.
To overcome this, a widely adapted approach is to rely on purely synthetic datasets \cite{fonder2019midair,wang2020tartanair,roberts2021hypersim}, which might not accurately represent the real world.

The authors of \cite{yao2020blendedmvs} propose a hybrid approach, reconstructing a 3D mesh from a set of images to render synthetic images and ground truth depth, which are blended with real images to create a realistic dataset.
Yet, the images are limited to those used for the reconstruction, which restricts the dataset diversity.
The authors of \cite{marelli2023enrich} propose a dataset for \ac{SfM} in built environments, using real-world 3D meshes to render synthetic viewpoints, effectively enriching the dataset with diverse perspectives.
Despite these efforts, a gap remains in datasets for \ac{MDE} and \ac{DC} that specifically target the field robotics domain.

\begin{figure*}[t]
    \centering
    \includegraphics[width=\textwidth]{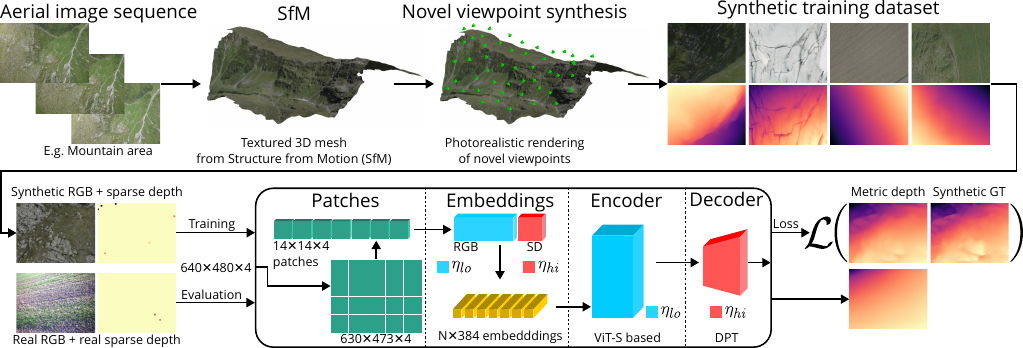}
    \caption{From an aerial image sequence, using \ac{SfM}, we obtain a textured 3D mesh of the area. With randomly sampled camera poses, we render synthetic RGB and depth ground truth for training. The input to the model is a four channel-image consisting of RGB and sparse depth (SD) information. The model operates on patches, which are transformed into an embedding vector, and through a \ac{ViT} and a \ac{DPT} decoder, we predict metric dense depth maps.}
    \label{fig:blockdiagram}
\end{figure*}
\section{Method}
In this section, we describe our method in detail.
Our primary objective is to develop a real-time capable metric \ac{DC} method tailored to unseen field robotics environments.
The architecture builds upon the existing \ac{SOTA} \ac{MDE} model, \ac{DAV2} \cite{yang2024depth_anything_v2}, which originally predicts affine-invariant depth.
We modify the convolutional layers of the pretrained encoder to accommodate a fourth input channel that represents sparse depth measurements.
The newly added random weights in these layers are trained from scratch, while the pretrained weights serve as initialization for the remaining model parameters.
We retrain the model on synthetic datasets derived from real-world textured 3D meshes, thereby enabling the incorporation of sparse depth measurements.
\Cref{fig:blockdiagram} provides an overview of the method.
In the following, we detail the training dataset generation, model architecture, loss function, and the synthetic depth measurement generation used during training.

\subsection{Synthetic Training Dataset Generation}
\label{sec:training_dataset}
\begin{figure}
    \centering
    \includegraphics[width=\columnwidth]{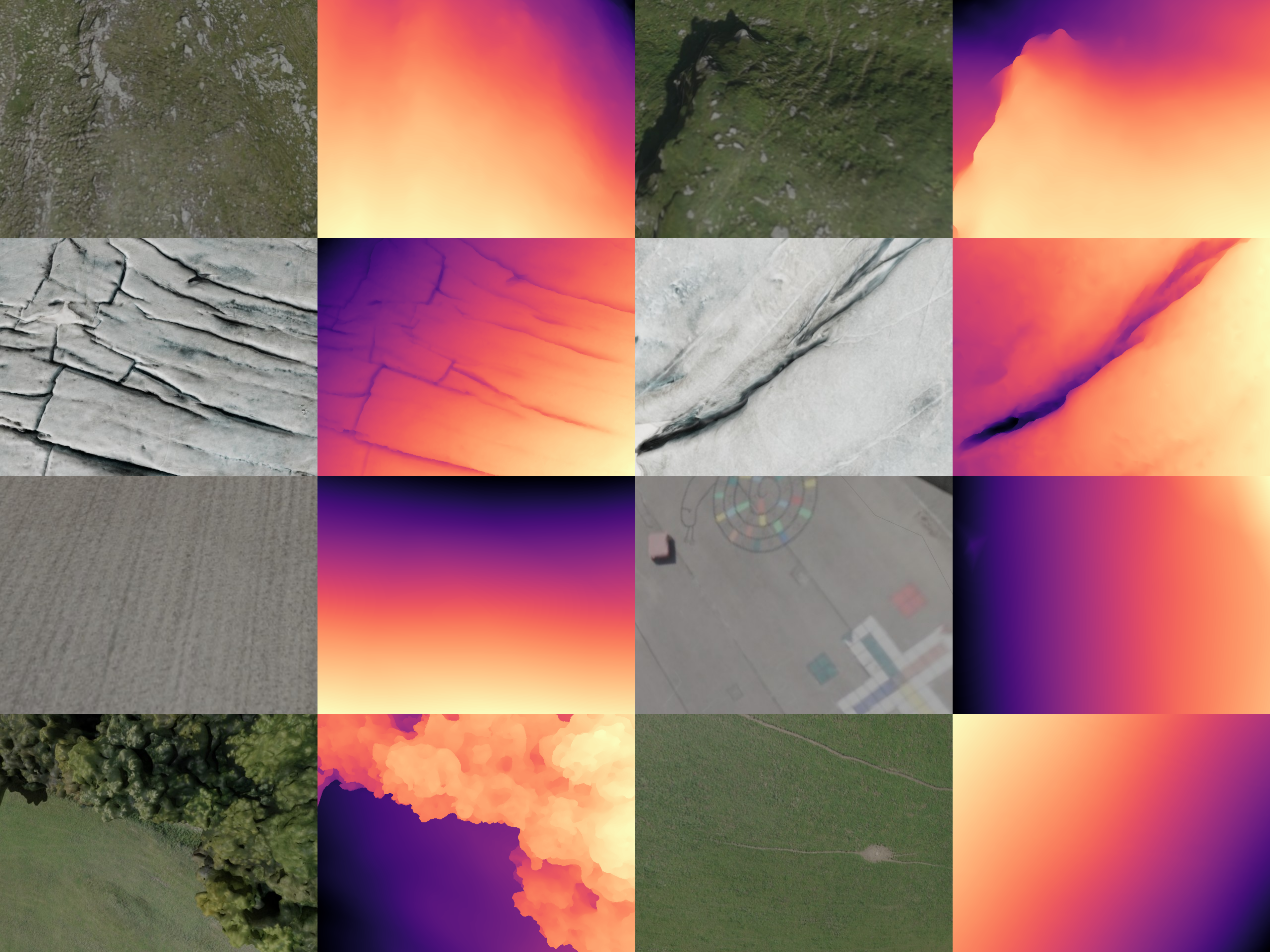}
    \caption{From top to bottom row, samples of the Mountain area, Rhône glacier, Road corridor, and Rural area synthetic training datasets. Each pair of columns shows an RGB image and its corresponding metric depth ground truth.}
    \label{fig:dataset_overview}
\end{figure}

The source data for the synthetic training datasets consists of multiple sequences of high-resolution RGB images captured from nadir cameras onboard \acp{UAV}.
Pix4D, a commercially available \ac{SfM} and photogrammetry software, is used to process the images and obtain a high-resolution, dense metric 3D mesh of the mapped region.
Four specific locations are selected to obtain four relevant datasets: A mountaineous area, a high-altitude alpine glacier, a road corridor, and a rural farming area.
These locations cover challenging aspects of field robotics such as self-similar texture in natural environments, a wide range of scale, and few human-built structures.
Such areas are not prominently represented in existing datasets, further justifying their selection.
For the dataset creation, the textured 3D mesh is processed in Blender, an open-source modelling and rendering software.
The goal is to generate a diverse synthetic RGB and ground truth depth dataset for training.
To achieve this, we randomly generate a set of $N_{frames}$ camera poses within the extent of the mesh in Blender.
The horizontal position of the camera pose is sampled within the mesh limits, while the vertical position (along gravity) is sampled within $\left[z_{min},z_{max}\right]$ above the mesh surface.
The camera's attitude along $x$, and $y$ is uniformly sampled within $\left[-\theta_{x,y}, \theta_{x,y}\right]$, while the rotation around the vertical axis is uniformly sampled over the full rotation.
Additionally, Blender's sun simulation plugin is used as the main light source, randomly sampling a time of day within $\left[t_{min}, t_{max}\right]$, simulating different intensities and shadows based on the mesh.
Blender then renders each camera pose, resulting in a set of high-quality, photorealistic input images and the corresponding metric depth ground truth map.
\cref{tab:ds_params} summarizes all parameters used for the dataset generation.
Following this approach, the synthetic training datasets for Mountain area, Rhône glacier, Road corridor and Rural area, shown in \cref{fig:dataset_overview}, are generated.
\renewcommand{\thetable}{\arabic{table}}
\captionsetup[table]{labelformat=simple, labelsep=colon, name=Tab.}
\begin{table}[t]
    \centering
    \caption{Generation parameters for the four synthetic training datasets.}
    \label{tab:ds_params}
    \begin{tabular}{lc}
        \toprule
        Generation parameter & Value \\
        \midrule
        Photogrammetry Rhône glacier            & DJI Mavic Pro \SI{12}{MP}\\
        Photogrammetry all others               & WingtraOne, Sony RX1RII \SI{42}{MP}\\
        \midrule
        $N_{frames}$                     & $4\times10,000$\\
        $\left[z_{min},z_{max}\right]$   & $\left[1, 51\right]$\,\unit{\metre}\\
        $\theta_{x,y}$                   & $22.5\,^\circ$\\
        $\left[t_{min}, t_{max}\right]$  & [10:00, 16:00]\\
        Dataset resolution               & $640\times480$\,\unit{px}\\
        \bottomrule
    \end{tabular}
\end{table}

\subsection{Architecture}
\label{sec:architecture}
The proposed \acf{DC} architecture uses the pretrained weights of the \ac{SOTA} \ac{MDE} model \acf{DAV2} \cite{yang2024depth_anything_v2}, originally designed for three-channel RGB images and affine-invariant depth prediction.
\ac{DAV2} is available in the \ac{ViT}-S variant, balancing between performance and inference speed.
Additionally, \ac{DAV2} is trained purely on synthetic or pseudo-labelled data, which fits well with our approach of using synthetic data for training \cite{yang2024depth_anything_v2}.

In our \ac{DC} approach, we encode the sparse depth measurements as a fourth input channel.
Below, we present the key modifications made to the model architecture to incorporate the fourth channel, followed by a detailed description of its representation.

\ac{DAV2} operates on $14\times14$ pixel patches of the input image, which are transformed into an embedding vector by a 2D convolutional layer.
We base our approach on the smallest \ac{ViT}-S \cite{touvron2021training} variant of \cite{yang2024depth_anything_v2} to minimize the latency for field deployments.
The embedding dimension of this variant is $384$, so the pretrained patch embedding 2D convolution weight has a shape of $384\times3\times14\times14$.
Since our approach uses four input channels, we concatenate this weight along the channel dimension with randomly initialized weights, forming a new tensor of shape $384\times4\times14\times14$.
The fourth channel provides sparse depth information at the pixel coordinates in the input image frame.
This modification enables the network to process the fourth channel while still using the initial pretrained weights as a starting point for the optimization.

Following \ac{DAV2}, we predict inverse (metric) depth and, for different cameras, use the canonical image space, as detailed in \cite{hu2024metric3dv2}.
For each sparse depth measurement $\mathbf{d}$, we find the corresponding $14\times14$ pixel patch in the input image and assign a transformed depth value $\mathbf{d}_{s,c,i}$ to the full fourth channel of the patch (visualized in \cref{fig:blockdiagram,fig:sparse_depth}).
The transformation for the sparse depth combines the canonical transformation ($\mathbf{d}_{c}$) from the camera's focal length $f$ to a canonical focal length $f_c$, inverts the depth ($\mathbf{d}_{c,i}$), and scales the normalized output ($\mathbf{d}_{s,c,i}$) to support the minimum depth $d_{min}$, since the network outputs values in $\left[0,1\right]$:
\begin{equation}
\begin{array}{ll}
    \mathbf{d}_{c}     = \frac{f_c}{f}\mathbf{d} & \mathbf{d}_{c,i}   = \frac{1}{\mathbf{d}_{c}} \\
    \mathbf{d}_{s,c,i} = d_{min}\mathbf{d}_{c,i}  & \\
\end{array}
\label{eq:canonical_depth}
\end{equation}

\subsection{Training Loss}
\label{sec:training_loss}
Since the prediction target is inverse metric depth, the loss is computed with respect to the inverse metric ground truth depth, transformed analogously with \cref{eq:canonical_depth}, from the synthetic training dataset as described in \cref{sec:training_dataset}.
The main loss function is the scale-invariant loss function proposed in \cite{eigen2014depth}:
\begin{equation}
\begin{aligned}
    \mathcal{L}_{si}(\mathbf{\hat{d}}, \mathbf{d}) = \frac{1}{N}\sum_{i=0}^{N-1} r_i^2 - \frac{\lambda_{si}}{N^2}\left(\sum_{i=0}^{N-1} r_i\right)^2\\
    \text{where } r_i = \log \hat{d}_i - \log d_i
\end{aligned}
\label{eq:loss_si}
\end{equation}
for all $N$ pixels in the depth prediction $\mathbf{\hat{d}}$ and the ground truth depth $\mathbf{d}$.
We utilize $\lambda_{si}=0.5$, a common choice to balance metric and ordinal correctness \cite{eigen2014depth}.
Additionally, we add a loss term, $\mathcal{L}_{grad}$ multiplied by a scaling factor $\lambda_{grad}$, for matching gradients between the prediction and the ground truth, as proposed in \cite{ranftl2022towards}.
This is especially effective for thin structures and sharp edges in combination with synthetic ground truths, as described in \cite{yang2024depth_anything_v2}.
The total loss function is then defined as $\mathcal{L}=\mathcal{L}_{si} + \lambda_{grad}\cdot\mathcal{L}_{grad}$.

\subsection{Synthetic Depth Measurements at Training Time}
\label{sec:sparse_prior}
\begin{figure}[h]
    \centering
    \includegraphics[width=\columnwidth]{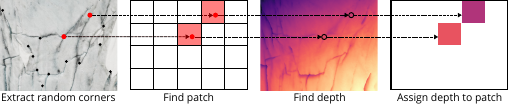}
    \caption{At each training step, random corners are sampled from the image (e.g. two corners here), and the corresponding patches are assigned depth values based on the ground truth. The patches are enlarged in this visualization.}
    \label{fig:sparse_depth}
\end{figure}

At training time, synthetic depth measurements are dynamically derived from the ground truth depth map to simulate the type of measurements typically available in field robotics applications.
The method detects simple and efficient corner features \cite{shi1994good} in the rendered RGB image and randomly samples between $\left[N_{d,min}, N_{d,max}\right]$ features as input.
The range is selected to roughly match the expected number of depth points in extremely sparse real-world scenarios.
This approach significantly increases the variability of the dataset, as the number of points and their distribution are different for each training step.
The generation process is illustrated in \cref{fig:sparse_depth}.

In addition, we add noise to the sparse synthetic depth points to increase the robustness to real-world sensor noise.
The noise is activated with probability $p_{noise}$ and takes the structure of a multiplicative noise on the depth value of the ground truth.
The multiplication factor is uniformly sampled between $\left[n_{low}, n_{high}\right]$.

\section{Experiments}
The experiments compare our approach against nine \ac{SOTA} baselines on five real-world field robotics datasets.
In particular, this section describes the characteristics, acquisition, and generation process of these datasets, and presents the training process of our model, specific details of the baseline works, and the evaluation process.

\subsection{Evaluation Datasets}
\label{subsection:evaluation_datasets}
The proposed approach is evaluated on five real-world field robotics datasets, all unseen during training: \acf{IH}, \acf{FF}, \acf{GF}, \acf{BT} and \acf{UF}.
These datasets represent diverse and challenging field robotics environments, including indoor, outdoor, underwater and aerial scenarios.
The first three datasets, obtained from dedicated field campaigns, are released with this paper; the remaining two are publicly available.
Specifically, the datasets \ac{IH} and \ac{FF} were collected using a hand-held rig in an industrial hall and on a farm field in Switzerland.
The \ac{GF} dataset was collected during a \ac{UAV} flight over the Rhône glacier in Switzerland.
The publicly available \ac{BT} and \ac{UF} datasets were collected in an aerated ship ballast tank \cite{ntnuarl2024ballasttank} and in the port of a fjord in Norway \cite{singh2024online}.
This paper also evaluates our approach on different sensor modalities: RGB or grayscale images and sparse depth measurements from a \ac{FMCW} radar, as well as sparse visual landmarks from ReAQROVIO \cite{singh2024online}.
The number of measurements from this modality is comparable to that from the radar sensors (see \cref{tab:sparsity_summary}).

\Cref{fig:introduction} shows representative example frames of the datasets with the sparse depth measurements overlaid in the RGB images.
An overview of the dataset properties is provided in \cref{tab:sparsity_summary}.

\renewcommand{\thetable}{\arabic{table}}
\captionsetup[table]{labelformat=simple, labelsep=colon, name=Tab.}
\begin{table}[b]
\centering
\caption{Overview of the real-world evaluation datasets: \acf{IH}, \acf{FF}, \acf{GF}, \acf{BT} and \acf{UF}. The column Frames refers to the number of frames in each dataset, while the column Measurements indicates the average number of depth measurements per frame and the column GT shows the average percentage of ground truth depth available.}
\label{tab:sparsity_summary}
\sisetup{detect-all}
\NewDocumentCommand{\B}{}{\fontseries{b}\selectfont}
    \begin{tabular}{@{}l l c c c@{}}
        \toprule
        Dataset & Sensors & {Frames} & {Measure-} & {GT (\%)}\\
        & & & {ments} &\\
        \midrule
        \ac{IH}  & \multirow{3}{*}{\shortstack{FLIR FFY-U3-16S2C-S\\+ TI AWR1843AOP}} & 289 & 1.7 & 9.1 \\
        \ac{FF}  &                                   & 321 & 2.5 & 11.2 \\
        \ac{GF}  &                                   & 349 & 2.6 & 0.6 \\
        \midrule
        \ac{BT} \cite{ntnuarl2024ballasttank} & Gray + TI IWR6843AOP & 299 & 4.3 & 0.7 \\
        \ac{UF} \cite{singh2024online} & Gray + visual landmarks     & 312 & 4.3 & 7.4 \\
        \bottomrule
    \end{tabular}
\end{table}

The \ac{IH} dataset is characterized by a dark, low light environment, with a repeating floor pattern.
The \ac{FF} dataset shows primarily vegetation and soil, with the absence of distinct human-built scale cues.
The \ac{GF} dataset is characterized by a dominant nadir view of the glacier, with snow, ice, crevasses and water streams.
The flight spans from zero to approximately \SI{30}{\metre} altitude, showing self-similar patterns in the ice and crevasses.
The \ac{BT} dataset is defined by a metal environment, captured in grayscale, with repeating patterns, low light and reflections, and some absolute scale features, such as manholes and ladders.
The \ac{UF} dataset represents the challenges of underwater perception, including limited visibility due to turbidity, overexposure from sunlight and suspended particles.
The dominant features are a pier wall and the seabed.

The datasets \ac{GF} and \ac{BT} were collected with Ouster OS1 and OS0-64 \ac{LiDAR} sensors, which are used as ground truth after reprojecting the point clouds to the image frame using known calibrations.
For the remaining datasets without a \ac{LiDAR}, \ac{SfM} is employed to reconstruct the scene and obtain a sparse ground truth point cloud.
For \ac{IH} and \ac{FF}, Pix4D and a DJI Mini 2 were used to collect \SI{12}{MP} nadir images of the same area.
In \ac{FF}, the geolocation metadata of the nadir images provides metric scale to the reconstruction.
In \ac{IH}, the dimensions of known objects in the scene are measured to provide metric scale.
In the \ac{UF} dataset, four cameras arranged in a static rig configuration with known calibrations are used in Colmap \cite{schoenberger2016sfm} to obtain a sparse metric point cloud.
Using the \ac{LiDAR} or \ac{SfM} pointclouds and poses, the sparse ground truth depth for each frame in the evaluation datasets is created.
Similarly, all sparse sensors (see \cref{tab:sparsity_summary}) return 3D pointclouds, which have been reprojected to the image frame using known calibrations.
As in \cref{sec:sparse_prior}, we assign the sparse depth value to the full patch corresponding to the pixel location of the measurement.
The same depth value transformation as in \cref{eq:canonical_depth} is employed.

\renewcommand{\thetable}{\arabic{table}}
\captionsetup[table]{labelformat=simple, labelsep=colon, name=Tab.}
\begin{table}[b]
    \centering
    \caption{Hyperparameters used for training our MDE network.}
    \label{tab:hyperparameters}
    \begin{tabular}{lc}
        \toprule
        Hyperparameter & Value \\
        \midrule
        Steps, batch size                & 14k, 64\\
        $(\eta_{hi}, \eta_{lo})$         & $(5\times10^{-5}, 5\times10^{-6})$\\
        $(\lambda_{si}, \lambda_{grad})$ & $(0.5, 0.5)$\\
        Optimizer, $(\beta_1, \beta_2)$  & AdamW \cite{loshchilov2018decoupled}, $(0.9, 0.999)$\\
        Weight decay, Gradient clipping  & $0.1$, max. norm $1.0$\\
        LR scheduler                     & Linear warmup until 1k steps\\
                                         & Constant until 2.5k steps\\
                                         & Cosine \cite{loshchilov2017sgdr} to 0.1, afterwards\\
        \midrule
        Color jitter prob., intensity          & \SI{80}{\percent}, [0.5, 1.5]\\
        Gamma prob., intensity                 & \SI{100}{\percent}, [0.5, 1.5]\\
        Horizontal flip prob., grayscale prob. & \SI{50}{\percent}, \SI{20}{\percent}\\
        Gaussian blur prob., $\sigma$          & \SI{20}{\percent}, [0.1, 2.0]\\
        \midrule
        $f_c$                                  & 900\\
        $(d_{min}, d_{max})$                   & (0.5, 80)\\
        $(N_{d,min}, N_{d,max})$               & (1, 10)\\
        \bottomrule
    \end{tabular}
\end{table}


\renewcommand{\thetable}{\arabic{table}}
\captionsetup[table]{labelformat=simple, labelsep=colon, name=Tab.}
\begin{table*}[ht!]
\centering
\caption{Overview of the performance of the baseline methods and our proposed approach. The table is sorted by the last column, the average rank. We highlight the best, second-best and third-best values with \colorbox{rankone}{first}, \colorbox{ranktwo}{second}, and \colorbox{rankthree}{third} place colors. The \ac{DC} column indicates whether the method is a depth completion method.}
\label{tab:results}
\sisetup{detect-all}
\def\Decimal{.000}
\begin{tabular}{
@{}
l  
c  
S[table-format=2.3]
S[table-format=2.3]
S[table-format=2.3]
S[table-format=2.3]
S[table-format=2.3]
S[table-format=2.3]
S[table-format=2.3]
S[table-format=2.3]
S[table-format=2.3]
S[table-format=2.3]
S[table-format=1.2]
@{}
}
\toprule
 & & \multicolumn{2}{c}{Industrial hall} & \multicolumn{2}{c}{Farm field} & \multicolumn{2}{c}{Glacier flight} & \multicolumn{2}{c}{Ballast tank} & \multicolumn{2}{c}{Underwater fjord} & \\
\cmidrule(lr){3-4} \cmidrule(lr){5-6} \cmidrule(lr){7-8} \cmidrule(lr){9-10} \cmidrule(lr){11-12}
Models & DC & {MAE} & {RMSE} & {MAE} & {RMSE} & {MAE} & {RMSE} & {MAE} & {RMSE} & {MAE} & {RMSE} & {Avg. rank}\\
\midrule
DepthPrompting (NYU)  \cite{park2024depthprompting} & \ding{51} & 4.974 & 5.140 & 7.198 & 7.606 & 7.175 & 7.232 & 1.530 & 1.768 & 4.221 & 4.765 & 8.50 \\
G2-MonoDepth \cite{wang2024g2monodepth} & \ding{51} & 4.725 & 5.096 & 4.656 & 5.185 & 5.099 & 5.491 & 3.327 & 3.609 & 4.747 & 5.281 & 8.10 \\
DAV2 \cite{yang2024depth_anything_v2} + LS & \ding{51} & 4.239 & 4.883 & 11.877 & 13.445 & 1.163 & 1.466 & 23.669 & 33.143 & \cellcolor{rankthree}2.000 & \cellcolor{rankthree}2.666 & 6.90 \\
Unik3D \cite{piccinelli2025unik3d} & \ding{55} & 2.611 & 2.738 & 3.552 & 3.724 & 8.018 & 8.031 & \cellcolor{ranktwo}0.821 & \cellcolor{ranktwo}1.018 & 2.700 & 6.085 & 6.40 \\
Depth Pro \cite{bochkovskii2025depthpro} & \ding{55} & \cellcolor{rankthree}1.728 & \cellcolor{rankthree}1.971 & 6.566 & 13.450 & 6.877 & 6.883 & 1.038 & 1.229 & \cellcolor{ranktwo}1.986 & \cellcolor{ranktwo}2.314 & 5.20 \\
Metric3DV2 \cite{hu2024metric3dv2} & \ding{55} & 2.156 & 2.256 & 4.185 & 4.465 & 6.393 & 6.404 & \cellcolor{rankthree}0.883 & \cellcolor{rankthree}1.087 & 2.346 & 3.088 & 5.20 \\
DepthPrompting (KITTI) \cite{park2024depthprompting} & \ding{51} & \cellcolor{ranktwo}1.273 & \cellcolor{ranktwo}1.591 & \cellcolor{rankthree}2.768 & \cellcolor{rankthree}3.250 & 4.907 & 5.280 & 2.117 & 2.427 & 3.944 & 4.340 & 5.10 \\
Marigold-DC \cite{viola2025marigolddc} & \ding{51} & 2.098 & 2.530 & 3.461 & 4.122 & \cellcolor{rankone}0.302 & \cellcolor{rankone}0.369 & 1.497 & 1.845 & 2.485 & 3.188 & \cellcolor{rankthree}\textbf{4.70} \\
OMNI-DC \cite{zuo2025omni} & \ding{51} & 1.977 & 2.335 & \cellcolor{ranktwo}2.247 & \cellcolor{ranktwo}2.802 & \cellcolor{rankthree}0.840 & \cellcolor{rankthree}1.026 & 1.272 & 1.619 & 2.304 & 2.987 & \cellcolor{ranktwo}\textbf{3.70} \\
Ours & \ding{51} & \cellcolor{rankone}0.930 & \cellcolor{rankone}1.294 & \cellcolor{rankone}1.155 & \cellcolor{rankone}1.532 & \cellcolor{ranktwo}0.534 & \cellcolor{ranktwo}0.620 & \cellcolor{rankone}0.734 & \cellcolor{rankone}0.979 & \cellcolor{rankone}1.523 & \cellcolor{rankone}2.129 & \cellcolor{rankone}\textbf{1.20} \\

\bottomrule
\end{tabular}
\end{table*}

\subsection{Training}
\cref{tab:hyperparameters} summarizes the hyperparameters used for our training approach.
We initialize training with the \ac{ViT}-S weights of \ac{DAV2} \cite{yang2024depth_anything_v2}, pretrained for affine-invariant depth, and through our loss function, retrained for metric depth prediction.
The concatenated random weights (see \cref{sec:architecture}) are trained with a high learning rate of $\eta_{hi}$.
The \ac{DPT} \cite{ranftl2021vision} decoder is also trained with a high learning rate, while the \ac{ViT} encoder is trained with a low learning rate $\eta_{lo}$.

The model is trained purely on synthetic training datasets (as introduced in \cref{sec:training_dataset}).
The training data is completely separate from the real-world evaluation datasets.
In addition to the datasets shown in \cref{tab:ds_params}, we also use $10k$ frames from each of the synthetic datasets Hypersim \cite{roberts2021hypersim} and Mid-Air \cite{fonder2019midair} for increased diversity.
We employ standard data augmentation techniques, namely color jittering, gamma changes, flipping, grayscale conversion and Gaussian blurring.
The parameters for the augmentations are show in \cref{tab:hyperparameters} as well.

\subsection{Baselines}
\label{subsection:baseline_works}
The baseline methods are selected based on their reported performance in the literature, with a focus on generalization capabilities, as we evaluate on datasets unseen during training.
We include \ac{MDE} methods because they demonstrate competitive performance compared to \ac{DC} methods in certain scenarios.
While real-time capabilities for diffusion-based methods are a concern in field robotics, we include them due to their \ac{SOTA} performance and future potential.
For all methods relying on \ac{ViT} architectures, the small \ac{ViT}-S variant \cite{touvron2021training} is used, if supported.
The Depth Pro \cite{bochkovskii2025depthpro} baseline uses a \ac{ViT}-L encoder, as a \ac{ViT}-S version is not released.
Following the proposal in \cite{viola2025marigolddc}, we perform a \ac{LS} fit on all sparse depth measurements to lift the affine-invariant prediction of \ac{DAV2} to the metric domain.
This baseline is denoted as DAV2+LS.
Additionally, the Metric3DV2 \cite{hu2024metric3dv2} method requires knowledge on the camera intrinsics, which we provide.
DepthPrompting provides two variants, trained on NYU-v2 and KITTI \cite{park2024depthprompting}.
We evaluate both variants on our field robotics datasets.

All baselines, and our approach, are evaluated in the same manner and with the same data, using the publicly available pretrained weights.
All evaluation datasets are rescaled to an intermediate resolution of $640\times480$ pixels, and evaluated at this resolution.

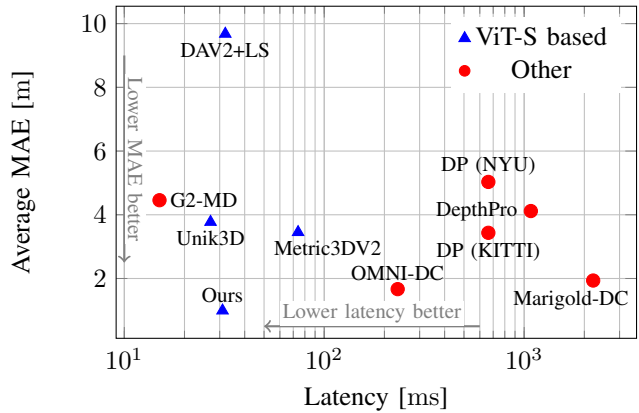
\begin{figure}[t]
    \begin{tikzpicture}
        \begin{axis}[
            xlabel={Latency [\unit{ms}]},
            ylabel={Average MAE [\unit{\metre}]},
            xmode=log,
            log basis x=10,
            ymajorgrids=true,
            xmajorgrids=true,
            grid=both,
            width=8.5cm,
            height=6cm,
            every node near coord/.append style={font=\footnotesize},
            legend style={at={(0.97,0.97)},anchor=north east,fill=white,draw=none},
        ]
        \addplot[only marks, mark=triangle*, color=blue, mark size=2.5pt] coordinates { (27,3.773) };
        \addlegendimage{only marks, mark=*, color=red}
        \addlegendentry{ViT-S based}
        \node[font=\footnotesize] at (axis cs:27,3.273) {Unik3D};
        \addplot[only marks, mark=triangle*, color=blue, mark size=2.5pt] coordinates { (31,0.991) };
        \node[font=\footnotesize] at (axis cs:31,1.491) {Ours};
        \addplot[only marks, mark=triangle*, color=blue, mark size=2.5pt] coordinates { (32,9.677) };
        \node[font=\footnotesize] at (axis cs:32,9.177) {DAV2+LS};
        \addplot[only marks, mark=triangle*, color=blue, mark size=2.5pt] coordinates { (74,3.452) };
        \node[font=\footnotesize] at (axis cs:104,2.952) {Metric3DV2};
        \addplot[only marks, mark=*, color=red, mark size=2.5pt] coordinates { (15,4.457) };
        \addlegendentry{Other}
        \node[font=\footnotesize] at (axis cs:25,4.457) {G2-MD};
        \addplot[only marks, mark=*, color=red, mark size=2.5pt] coordinates { (233,1.666) };
        \node[font=\footnotesize] at (axis cs:233,2.166) {OMNI-DC};
        \addplot[only marks, mark=*, color=red, mark size=2.5pt] coordinates { (662,3.434) };
        \node[font=\footnotesize] at (axis cs:662,2.834) {DP (KITTI)};
        \addplot[only marks, mark=*, color=red, mark size=2.5pt] coordinates { (662,5.031) };
        \node[font=\footnotesize] at (axis cs:662,5.531) {DP (NYU)};
        \addplot[only marks, mark=*, color=red, mark size=2.5pt] coordinates { (1080,4.117) };
        \node[font=\footnotesize] at (axis cs:580.0,4.117) {DepthPro};
        \addplot[only marks, mark=*, color=red, mark size=2.5pt] coordinates { (2214,1.936) };
        \node[font=\footnotesize] at (axis cs:1714,1.336) {Marigold-DC};

    \draw[->, thick, gray] (axis cs:600,0.5) -- (axis cs:50,0.5) node[midway, above, sloped, font=\footnotesize, fill=white, inner sep=1pt] {Lower latency better};
    \draw[->, thick, gray] (axis cs:10,9) -- (axis cs:10,2.5) node[midway, above, font=\footnotesize, fill=white, inner sep=1pt, rotate=-90] {Lower MAE better};
    \end{axis}
    \end{tikzpicture}
    \caption{Latency, defined as the total time required to process input and generate output on an Nvidia 3090 GPU, plotted against the average MAE across all experiments.}
    \label{fig:latency_vs_mae}
\end{figure}

\section{Results}
The obtained comparison results are presented in \cref{tab:results}.
The evaluation metrics are \ac{MAE} and \ac{RMSE}, following \cite{viola2025marigolddc}, and are averaged over all frames in each dataset.
The table is sorted by the average rank across all metrics and datasets.
Additionally, \cref{fig:introduction} provides qualitative results that illustrate the visual quality of the depth predictions.

At a high level, our approach achieves the highest average rank, demonstrating its effectiveness in adapting to unseen, real-world environments and noisy depth data.
In the individual experiments, our approach performs consistently, achieving the lowest error metrics except in the Glacier flight experiment.
The overall second- and third-ranked methods do not consistently achieve the second and third rank in each individual experiment.
These observations suggest that our approach is more robust, and the dedicated design choices in model architecture and training protocol are effective in these scenarios.

The Ballast tank experiment is particularly challenging, as the radar measurements are quite noisy due to the metal environment; this may explain the low performance of \ac{DAV2}+LS, where fitting the sparse depth data to the dense affine-invariant depth yields very high errors.
All \ac{DC} methods (except our method) are outperformed by the \ac{MDE} methods, further supporting this hypothesis.
Similarly, in the Underwater fjord experiment, the second-best method is a \ac{MDE} method.
This suggests that noisy and extremely sparse depth data can have adverse effects on the performance of \ac{DC} methods when treating the problem purely as a regression or in-painting task.

In the Glacier flight scenario, Marigold-DC \cite{viola2025marigolddc} achieves a lower \ac{MAE} and \ac{RMSE}.
This scenario is challenging due to the nadir-view perspective, which does not provide strong clues on the orientation of the glacier.
Marigold-DC appears to predict more planar depth maps that are aligned more accurately with the depth measurements, explaining the higher performance.

For practical deployment in the field, we highlight three key aspects: inference time, dataset size, and the use of synthetic data.
The second- and third-ranked methods are likely not suited for real-time applications without further optimisation (see \cref{fig:latency_vs_mae}).
We also conduct an additional inference time analysis of our method on a Nvidia Jetson AGX Orin platform for relative comparison between a desktop GPU and this embedded device.
With an inference time of \SI{53}{ms} on the embedded device, our approach is suitable for real-time onboard computation in many mobile robotics applications.
By partially re-using pretrained weights of \ac{DAV2}, our approach utilizes a relatively small dataset size ($60k$ ours versus $573k$ for OMNI-DC), and a short training time (approximately \SI{7}{hours} on a single Nvidia 3090 GPU) until convergence.
Lastly, training on a diverse synthetic dataset with simulated depth sensor data, all derived from real-world textured 3D meshes, suggests a promising direction, with our approach generalizing well to the real-world experiments.

\section{Conclusions}
This paper presents a novel approach for \acf{DC} with extremely sparse depth measurements, a common scenario in challenging field robotics applications.
The model enables the combination of a monocular camera and sparse depth sensors as a viable dense metric depth sensor modality.
Our approach achieves the highest average rank across all investigated unseen field robotics evaluation datasets, demonstrating the effectiveness of the proposed methodology across different scenarios while also keeping inference time low for real-time applicability.
In addition to a \ac{DC} method, this work also proposed a method for generating large amounts of synthetic training data for field robotics applications and releases relevant field robotic datasets.
This contribution is expected to significantly reduce manual data collection and simultaneously address challenges related to incomplete or noisy depth ground truth.
While the proposed method is specifically designed for extremely sparse depth measurements, it does not generalize well to very dense depth measurements due to the specialized depth input space and training process.
Addressing this limitation to support any depth sensor density could be explored in future work.

\printbibliography
\end{document}